# Synchronous Adversarial Feature Learning for LiDAR based Loop Closure Detection *

Peng Yin, Yuqing He, Lingyun Xu, Yan Peng, Jianda Han and Weiliang Xu.

*Abstract*— Loop Closure Detection (LCD) is the essential module in the simultaneous localization and mapping (SLAM) task. In the current appearance-based SLAM methods, the visual inputs are usually affected by illumination, appearance and viewpoints changes. Comparing to the visual inputs, with the active property, light detection and ranging (LiDAR) based point-cloud inputs are invariant to the illumination and appearance changes. In this paper, we extract 3D voxel maps and 2D top view maps from LiDAR inputs, and the former could capture the local geometry into a simplified 3D voxel format, the later could capture the local road structure into a 2D image format. However, the most challenge problem is to obtain efficient features from 3D and 2D maps to against the viewpoints difference. In this paper, we proposed a synchronous adversarial feature learning method for the LCD task, which could learn the higher level abstract features from different domains without any label data. To the best of our knowledge, this work is the first to extract multi-domain adversarial features for the LCD task in real time. To investigate the performance, we test the proposed method on the KITTI odometry dataset. The extensive experiments results show that, the proposed method could largely improve LCD accuracy even under huge viewpoints differences.

## I. INTRODUCTION

Loop Closure Detection (LCD) is the key module in the SLAM scheme of mobile robot [1–3]. LCD could enhance the robustness of SLAM algorithms in detecting when a robot has returned to a previous visited location after having discovered new terrains. Such detection makes it possible to increase the precision of the actual localization estimate and helps to resolve the global localization problem. Traditionally there exists two categories of SLAM approaches, metric based SLAM [4] [5] and appearance based SLAM [6] [7]. The metric-based SLAM methods usually rely on dense or semi-dense map reconstruction to achieve high accurate localization. While this method is impractical for the long-term navigation task, such as for the Autonomous Car traveling for thousands of miles or lifelong tended outdoor service robots. Instead, appearance based SLAM (or Topological SLAM [1]) only needs to extract the low dimension semantic features to describe the local map. The computation complexity and storage requirement of appearance-based SLAM methods are relative low than the metric-based ones, thus the former could be easily applied on normal robot systems.

However, visual based appearance inputs are usually effected by the *illumination* changes, season-to-season based *appearance* changes [6–9] and *viewpoints* differences. To extract stable LCD features from visual inputs under all conditions is intractable. SeqSLAM [7] [10] uses the most similar coherent sequence as the best match instead of calculating the most likely single pair match [6] [11]. Though this method could improve the LCD accuracy under variant conditions, it may still fail under extreme visual appearance difference. Most recently, deep neural networks (DNN) have been used for visual feature extraction in the SeqSLAM. Chen et.al [12] extracted the multi-layer DNN features for LCD task. The network model is pre-trained under the ImageNet 2012 dataset [13]. Sunderhauf et.al investigated the ability of different layers for LCD detection [14]. Lowry [15] proposed a Change Removal approach, and use the non-common geometry for the DNN feature extraction. Most recently, Chen et.al [16] enable the feature learning by transforming the DNNs into a scene classification task. The above DNN based feature extraction could only be applied for the visual inputs, and must be supported by additional data labels, such as the ImageNet [17], KITTI [18] or self-defined datasets [16].

An alternative to the visual images is the active 3D point-cloud from LiDAR inputs, which is inherently invariant to the *illumination* and *appearance* changes. The only thing that matters the LCD accuracy with LiDAR inputs, is the *viewpoints* differences. In this paper, we extract 3D voxel map and 2D top-view map from the raw point-cloud, and each kind of map has its own advantage in local map representation. As we can see in Figure 1, for the 3D voxel maps, it could capture the local geometry detail into a simplified 3D voxel format; for the 2D top-view maps, it could easily capture the local road structure into a 2D image format. Despite their property, extracting efficient features from 3D and 2D maps is intractable. Traditional handcraft features couldn't capture the global connections of local geometry details; thus, the LCD accuracy will reduce greatly under huge viewpoints difference. On the other hand, the 3D feature [19–21] extraction is time consuming, and could not be easily applied in the real time LCD task.

In this paper, we propose a novel synchronous adversarial feature learning (Sync-AFL), a kind of unsupervised learning method for LiDAR based LCD task.

*Research supported by Natural Science Foundation of China (No.U1608253, 61473282, 61403245).

Peng Yin and Lingyun Xu is with the State Key Laboratory of Robotics, Shenyang Institute of Automation, Chinese Academy of Sciences, Shenyang 110016, University of Chinese Academy of Sciences, Beijing 100049, Email: yinpeng, xulingyun@sia.cn.

Yuqing He, and Jianda Han are all with the State Key Laboratory of Robotics, Shenyang Institute of Automation, Chinese Academy of Sciences, Shenyang 110016, China. Email: heyuqing, jdhan@sia.cn.

Yan Peng is with the School of Mechatronic Engineering and Automation, Robotics, Shanghai, China. Email: pengyan@shu.edu.cn.

Weiliang Xu is with Department of Mechanical Engineering University of Auckland, New Zealand China. Email: p.xu@auckland.ac.nz.

The proposed method requires no label data and could extract the mixture semantic features from both 2D and 3D maps at real time. The contribution of this paper could be summarized as below:
- With the invariant property of LiDAR inputs for illumination and appearance changes, we extract 3D voxel maps and 2D top-view maps from the raw point-cloud for LCD detection. Each type of map has its own advantage in local map description.
- We proposed a novel Sync-AFL method, and this method enable the multi-type feature extraction at the real time. And the feature learning is proceed under a synchronous manner, which enforce the mixture feature extraction from the potential data distribution.
- We investigate the performance of the proposed method on the KITTI odometry datasets. And the results show that the LCD accuracy is significantly improved, even under huge viewpoints differences.

The paper is organized as follows: in Section II, we first provide the primary works which our work is based on; in Section III, we will mainly introduce the Sync-AFL based LCD method; Section IV demonstrates the experiment results on the KITTI odometry datasets; finally, Section V gives the conclusion and consideration for our future works.

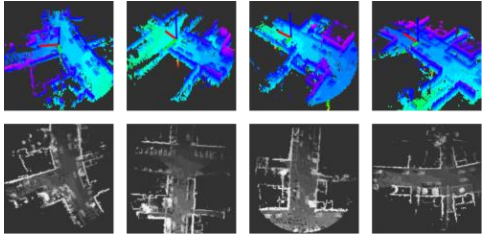

**Figure 1** The 3D voxel maps (first row) and 2D top-view maps (second row).

I. PRIMARY

Our Sync-AFL method is based on the bidirectional generative adversarial networks (BiGAN), so in this section, we will firstly review the BiGAN [22] based adversarial feature learning (AFL) method.

*A. Adversarial Feature Learning*

Recently, GANs [23] has gained much attention in represent learning [24], image generation [25] and unsupervised learning [22] [26]. As shown in Figure 2(a), the networks of GANs is combined with a decoder module and a discriminator module. The decoder module $De$ intend to generate the synthesis data $G\text{-}X$ from the random latent code $Z$ as real as possible, and the discriminator module $D$ is used to distinguish the synthesis data from the real ones as accurate as possible. With this two-player game, the synthesis data distribution could be pull to the real data distribution. Goodfellow [23] use the min-max value function $\min_{De}\max_{D} V(D, De)$ to achieve the above requirements, where the value function is obtained by,

$$V(D, De) = E_{x \sim P_X(x)}\left[\log D(X)\right] + E_{z \sim P_Z(z)}\left[\log\left(1 - D(De(z))\right)\right] \quad (1)$$

where $P_X$ and $P_Z$ is the data domain and latent code domain relatively.

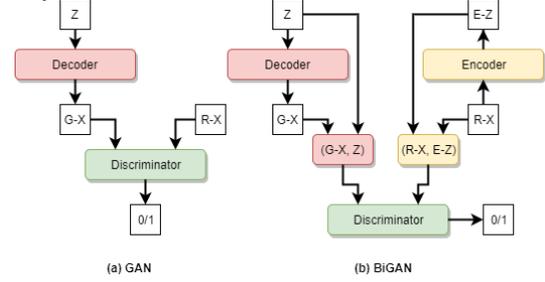

**Figure 2** The structure of DCGAN and BiGAN.

However, GANs itself could not directly obtain the latent codes from the real data domain, Donahue et.al [22] enable the latent code inference by proposing the BiGAN method as shown in Figure 2(b). This method adds extra encoder module $En$ to inference latent codes $E\text{-}Z$ from the real data $R\text{-}X$, and update the discriminator module to distinguish the mixture distribution of (data, latent code), instead of the single data domain. Then the min-max value function is updated into $\min_{De, En}\max_{D} V(D, De, En)$, where the value function in Equation 1 is updated into,

$$V(D, De, En)$$
$$= E_{x \sim P_x}\left[\log D(x, En(x))\right] + E_{z \sim P_z}\left[\log\left(1 - D(De(z), z)\right)\right]$$
$$= E_{x \sim P_x}\left[E_{z \sim P_{En}(\cdot|x)}\left[\log D(x, z)\right]\right] +$$
$$E_{z \sim P_z}\left[E_{x \sim P_{De}(\cdot|z)}\log\left(1 - D(x, z)\right)\right]$$
$$\quad (2)$$

where $P_{En}$ is the encoder mapping distribution from data domain to latent code domain, $P_{De}$ is the decoder mapping distribution from latent code domain to data domain. As proved by the Donahue [22], with fixed $En$ and $De$ modules, the value function $V(D, De, En)$ could reach its optimal with the optimal discriminator $D_J*$. And under the optimal discriminator $D_J*$, the value function $V(D_J, De, En)$ could be rewritten as [22],

$$C(De, En) = -2\log(2) + 2JSD(P_{EX} \| P_{GZ}) \quad (3)$$

where $P_{EX}$ is the mixture distribution of $(R\text{-}X, E\text{-}Z)$ and $P_{GZ}$ is the mixture distribution of $(G\text{-}X, Z)$. The optimal discriminator could capture the Jensen-Shannon divergence (JSD) between the mixture distribution $P_{EX}$ and $P_{GZ}$. Since JSD is always non-negative, so the value function could only reach its global optimal when $P_{EX}=P_{GZ}$. With the optimal encode $En$, we could easily extract the latent code from the real data by forward steps.

II. SYNC-AFL BASED LCD

As shown in the Figure 3, the proposed Sync-AFL based LCD framework is combined with four steps:
- Step1: 3D/2D map extraction
- Step2: Synchronous adversarial feature learning
- Step3: Real Time feature Inference
- Step4: Sequence Matching

In Step1, the 3D voxel maps and 2D top-view maps are generated with a dynamic octree mapping method. In Step2,

the adversarial networks is trained with the local 3D and 2D maps under a synchronous manner. In Step3, the mixture feature is extracted from both the 2D top-view maps and 3D voxel maps. Finally, in Step4, the best matches are obtained by using the sequence matching method. In this section, we will explain each step in detail.

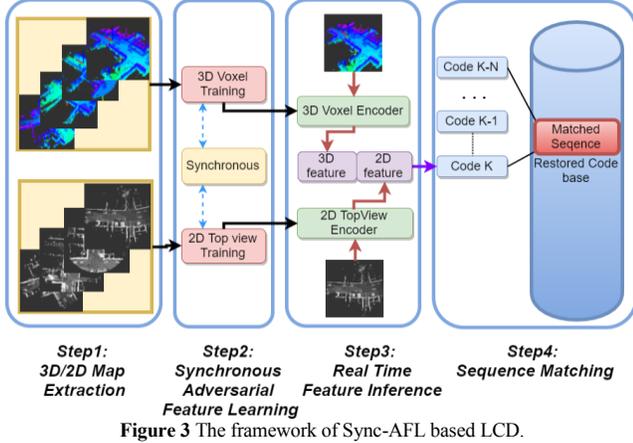

**Figure 3** The framework of Sync-AFL based LCD.

### A. 3D/2D Map Extraction

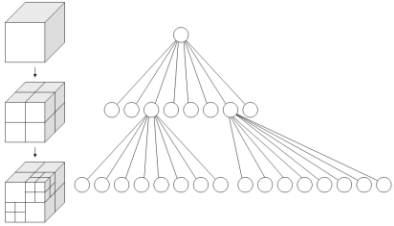

**Figure 4** Octree Structure

Since single LiDAR scan is too sparse, directly using single raw point-clouds could not capture the detail geometry. So in this paper, we use the Octomap [27] method to accumulate the sequence raw point-clouds. The Octomap is based on the octree structure, and the Octree is a tree-based 3D data structure as shown in Figure 4. From the root node, each node divides its space into eight sub nodes with eight equal sub spaces. The sub nodes continue to divide the space until the depth of node reach the given threshold. The occupancy of the leaf nodes are updated with a log-odds [28] approach. For the detail of the octree mapping, please refer to the original Octomap method [27].

In Octomap, the map is restored into a global static octree structure, and the mapping efficiency would be reduced significantly as the map growing into large scale. In this paper, we use the raw LiDAR point-clouds to construct the static octree map, but only keep the local octree nodes which are within a given distance to the robot (30 meters in this paper). This trick could enable the local 3D octree mapping with limited computation power and storage and guarantee a stable map generation in 7~10Hz. As shown in Figure 1, the 3D voxel maps are generated from the leaf node of local octree map, and the 2D top-view maps are generated by projecting the 3D voxel map onto the ground plane. The 2D top-view maps could capture the main structure of the road, and the 3D voxel maps contain the geometry detail into a simplified 3D voxel format.

### B. Synchronous Adversarial Feature Learning

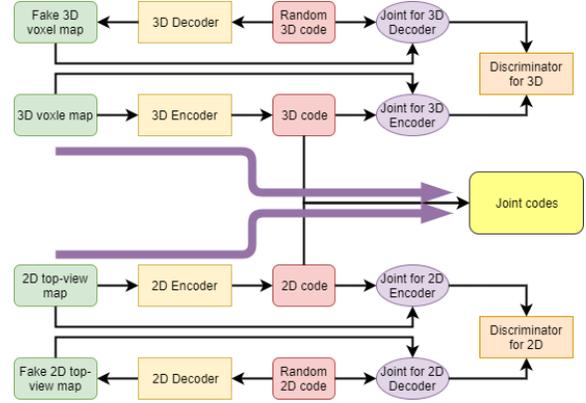

**Figure 5** Flowchart of the unsupervised feature training networks. This mixture framework is combined with 3D and 2D BiGAN networks.

As shown in Figure 5, in our Sync-AFL method, we use a dual-BiGAN, which combined two independent 2D and 3D BiGAN. For 2D top-view maps, the 2D BiGAN is based on 2D convolution/deconvolution operations. While for 3D voxel maps, the 3D BiGAN is based on 3D convolution/deconvolution operations. The mixture features of 3D/2D latent codes are obtained by a stitching operation. Though this method, we could inference the mixture features from both 3D and 2D maps at the same time, which could capture the LCD characteristics from different views.

Though we could train the 2D and 3D BiGAN separately, we could not guarantee the mixture 2D-3D feature could improve the overall LCD accuracy. This is caused by the model missing problem [29] [30] in the original GANs method, where the synthesis data could only focus on the part of real data distribution. When trained the 3D/2D BiGANs in separate branches, the synthesis 3D/2D data distributions may cover different parts of the real data distribution. Thus, the mixture 3D/2D features may get stocked into a confusing situation.

To enhance the 3D-2D feature connection, we use a synchronous training manner. The basic idea of this method is to reweighting the samples according to the *"unfamiliarity"* for the local scenes. Ideally, the more the network is familiar with a specific scenes, the more the discriminator will value the scene with a high score. Here, we measure the *unfamiliarity* of the scene based on the reciprocal of the average discrimination values,

$$U(M_{3D}, M_{2D}) = \left(\frac{D_{3D} + D_{2D}}{2}\right)^{-1} \quad (4)$$

where $D_{3D}$ and $D_{2D}$ are the discrimination values of 3D map $M_{3D}$ and 2D map $M_{2D}$. The lower $U$ means the 3D-2D BiGANs are familiar with the scene, while the higher $U$ means that at least one module is unfamiliar with the scene. The weighting of each sample scenes are labeled according to the unfamiliarity with a normalization function,

$$W_i = \frac{\exp(U_i)}{\sum_{j=1}^{N} \exp(U_j)} \quad (5)$$

The less valued real samples will be given higher weights, so such samples will play more important roles in the next epoch network updating. The whole algorithm is given in Algorithm 1. In each epoch, as shown in line 3~9, the 3D and 2D BiGANs are updated with the given 3D map $M_{3D}$ and 2D map $M_{2D}$, and calculate the sample's unfamiliarity. Then in line 10, the training samples are reweighted according to Equation 5.

| Algorithm 1 **Synchronous Training** |
|---|
| Inputs: $N$ samples of ($M_{3D}$, $M_{2D}$) |
| 0 Initialize pair weighting to $1/N$ |
| 1 For Epoch in Max-Epochs: |
| 2   Samples shuffle; |
| 3   For ($M_{3D}$, $M_{2D}$) in samples: |
| 4     3D BiGAN updating based on $M_{3D}$; |
| 5     Calculate $D(M_{3D}, En(M_{3D}))$ as $D_{3D}$; |
| 6     2D BiGAN updating based on $M_{2D}$; |
| 7     Calculate $D(M_{2D}, En(M_{2D}))$ as $D_{2D}$; |
| 8     Estimate the sample ($M_{3D}$, $M_{2D}$) unfamiliarity. |
| 9   End For |
| 10  Re-weightings N samples. |
| 11 End For |

By using this approach, the training procedure will continue updating the sample weightings, and focus the unfamiliar samples, thus finally improve the 3D-2D features connections on the real data distribution.

### C. Real Time Feature Inference

The feature inference step could achieved by applied the forwarding steps in 3D encoder and 2D encoder modules, this operation is easily to achieved with the support of normal GPU card or the embedded board Jetson TX1 or TX2. This property enables our method to imply on the traditional normal robots for the long term navigation task in real time. With the extracted mixture features, we could use Euclidean distance, cosine distance or other format distance to calculate their similarity. In this paper, we simply use the Euclidean distance,

$$Diff(v_i, v_j) = \|v_i - v_j\|^2 \quad (6)$$

where $v_i$ is the encoded latent-code from the frames.

## III. EXPERIMENT AND ANALYSIS

To investigate the performance of our method, the experiments are conducted on the KITTI odometry[1] datasets. Within this dataset, there are 22 LiDAR sequences and only the sequence 00~10 have the ground truth GPS location. Among these sequences, we extract the 3D voxel maps and 2D top-view maps from sequence 01~08 for the SAFL networks training. And use sequence 00, 09 and 10 for LCD accuracy testing. The SAFL networks is applied on a single NVidia Titan X card with 64G RAM. The 3D/2D map extraction is based on the robot operation system (ROS). Since the inherently invariant property of the LiDAR inputs to the appearance and illumination changes, the only thing that affect the LCD accuracy is the viewpoints differences. In practice, the local difference in pitch, roll and height are reduced by the log-odds based octree mapping step. For two map sequences, the major viewpoints differences come from the *translation difference* on the ground plane and the *Heading difference*.

To test the robustness of our proposed SAFL method in the LCD task, we generate different test sequences with $T_{\{Tt\}}\_R_{\{Rh\}}$ setting, where the translation is added with a two-dimensional random noise (amplitude is $T_t$ meters), and the heading angle is added with one-dimensional random noise (amplitude is $R_h$ radian). Since our method use the LiDAR inputs for the LCD task, so it will be meaningless to make the comparison with the appearance based methods. Here, the LCD is estimated under the same SeqSLAM based sequence matching framework

### A. Measurement Metrics

To measure the LCD accuracy of different methods, we make qualitative analysis with PRC (Precision-Recall curve) and AUC (area under the Receiver operating characteristic (ROC)); for the quantitative analysis, we use the recall at 100% perception in the PRC to measure LCD accuracy. Here, for the matched pairs, if the distance between ground truth position and estimated one is within $D_{thresh}$, then the pairs are regarded as true positive (TP), else will be regarded as false positive (FP); on the other side, the pairs erroneously discarded by the match score are regarded as false negative (FN), and the ones of no-matched pairs are regarded as the true negative (TN). Thus, the precision and recall are then obtained by,

$$\Pr ecision = \frac{TP}{TP+FP}$$
$$\operatorname{Re}call = \frac{TP}{TP+FN} \quad (7)$$

The AUC score is the size of covered ROC area, where the higher the score, the more accurate of the LCDs. The ROC curve is created by plotting the true positive rate (TPR) against the false positive rate (FPR) at various threshold settings, which are obtained by,

$$TPR = \frac{TP}{TP+FN}$$
$$FPR = \frac{FP}{TN+FP} \quad (8)$$

### B. Accuracy and efficiency analysis

In this section we first given the qualitative analysis with PRC and ROC index for each method. Figure 9 shows the PRC of the SAD based LCD accuracy in the original SeqSLAM method. We can see in the smaller viewpoints difference cases; the SAD features could still achieve a stable LCD accuracy. While for the higher viewpoints difference cases, the LCD accuracy is decreasing.

---
[1] http://www.cvlibs.net/datasets/kitti/eval_odometry.php

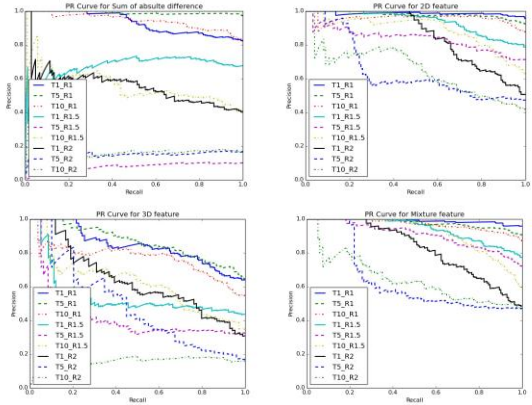

**Figure 6** Precision-Recall Curve under different noising sequences in different methods. The feature difference between two places is estimated by calculating the Euclidean difference.

Figure 6 shows the PRC of different methods. 2D top-view map features could significant improved than the SAD features. This explains the 2D adversarial features extracted from the 2D top-view maps can capture higher level connections of the local scenes, which could improve the robustness for the viewpoints changes. For 3D voxel map feature, though the results are still better than the SAD features, it is worse than the 2D features. This could be explained in two way: firstly, the 3D voxel maps is more difficult than the 2D top-view maps, so the 3D networks is more complex than 2D and the efficient 3D feature extraction is even more complex; secondly, in the KITTI odometry datasets, the 2D top-view has captured the major structure for efficient LCD detection, though 3D voxel map has the additional height dimension, the geometry in height may introduce limited benefit for LCD detection. Despite the 3D features is not better than 2D features, the mixture feature is even better than the 2D feature based sequence matching. It shows that, with our proposed synchronous training method, the mixture features could be better capture the higher representation for the local scenes.

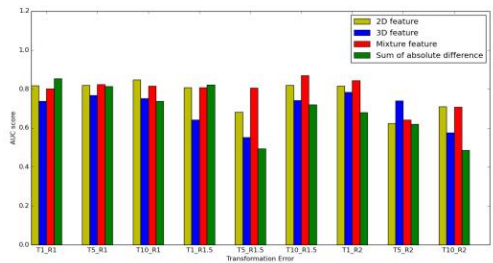

**Figure 7** The area under an ROC curve (AUC) score for each sequence matching method under different noising sequences.

TABLE I
RECALL AT 100% PRECISION

| Method | T1 R1 | T5 R1 | T10 R1 | T1 R1.5 | T5 R1.5 | T10 R1.5 |
|---|---|---|---|---|---|---|
| *SAD* | 28.3% | **44.5%** | 12.4% | 1.3% | 0.8% | 0.9% |
| *2D* | 51.9% | 19.8% | 35.7% | 26.5% | 3.8% | 9.1% |
| *3D* | 21.6% | 15.5% | 3.8% | 5.2% | 5.3% | 8.1% |
| *Mix* | **52.2%** | 43.6% | **43.8%** | **49.7%** | **18.2%** | **20.5%** |

TABLE III
FEATURE INFERENCE TIME PER FRAME (MILLISECOND)

| Method | T1 R1 | T5 R1 | T10 R1 | T1 R1.5 | T5 R1.5 | T10 R1.5 |
|---|---|---|---|---|---|---|
| *SAD* | 3.2 | 3.3 | 3.3 | 3.3 | 3.2 | 3.3 |
| *2D* | 18.4 | 16.9 | 17.6 | 17.7 | 17.7 | 15.6 |
| *3D* | 96.2 | 105.7 | 96.4 | 94.8 | 95.0 | 92.5 |
| *Mix* | 110.6 | 117.3 | 110.9 | 109.6 | 110.4 | 104.3 |

For a more intuitive view, Figure 7 shows the AUC index of different methods. The mixture feature based sequence matching is better in most viewpoints differences cases. Finally, TABLE II gives the quantitative analysis with recall rate at 100% precision. In the case of *T10_R1*, the recall rate of LCD task base on mixture features is 43.8%, which is 353% than the SAD feature, 122% than the 2D based features and 11 times than the 3D features.

### C. Storage and Runtime Analysis Data Reconstruction

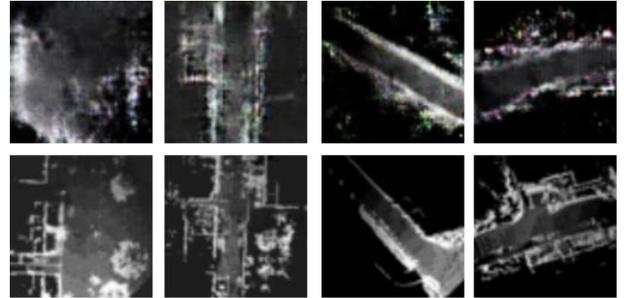

**Figure 8** 2D top-view maps reconstruction. The bottom row is the original frames, the top row is the cycle reconstructed ones.

Since our method is based on the BiGAN, the data reconstruction from real data $X$ into $De(En(X))$ as shown in Figure 8. The better the reconstruction results, the better ability for the encoder module to capture he geometry detail from the real data. For the simple road structure case, such as straight road or simple square space, the data reconstruction is quite well.

### D. Storage and Runtime Analysis

The mixture features are saved as a 1024 vector in the float32 format, so each code is occupied for 1024*4B=4KB. If we generate the code at 5Hz, after 24 hours running, the storage requirement for saving all the latent codes is only about 5*60*60*24*4KB~1.65GB. For the runtime analysis of feature inference, the average feature inference time is shown in Table III. As we can see, the mixture features 2D feature per frame could be estimated around 20ms, 3D features around in 110ms.

Thus, the proposed method could be easily plugged on the any kinds of mobile robots for the real time long term navigation task, with relative small computation power and storage requirement.

### IV. CONCLUSION

In this paper we propose a novel synchronous adversarial feature learning method for LiDAR based Loop closure detection. Based on the LiDAR inputs, we extract 3D voxel map and 2D top-view map to describe the local scene in difference angels. And our proposed adversarial feature learning method could extract the mixture features from both 3D and 2D maps at the same time. With the synchronous training scheme, the proposed mixture features could capture the potential data distribution as much as possible, and could improve the robustness for the viewpoints difference. The proposed method could be easily applied on normal robots for the long term LCD task, without the required for higher computation power and larger storage space. In our future work, we will test this method in the more complex 3D environment, such as in the terrain area, forest and ruins, where the 3D features could be very necessary for the loop closure detection.